\documentclass{article}

\usepackage{arxiv}
\usepackage{float}
\usepackage[table,xcdraw]{xcolor}
\usepackage{amsmath,amssymb}
\usepackage{graphicx}
\usepackage{caption}
\usepackage{subcaption}
\usepackage{algorithm,algpseudocode}
\usepackage[utf8]{inputenc} % allow utf-8 input
\usepackage[T1]{fontenc}    % use 8-bit T1 fonts
\usepackage{hyperref}       % hyperlinks
\usepackage{url}            % simple URL typesetting
\usepackage{booktabs}       % professional-quality tables
\usepackage{amsfonts}       % blackboard math symbols
\usepackage{nicefrac}       % compact symbols for 1/2, etc.
\usepackage{microtype}      % microtypography
\usepackage{lipsum}
\DeclareMathOperator*{\argmin}{argmin}

\title{Noise-Aware Texture-Preserving Low-Light Enhancement}

\author{
  Zohreh Azizi \\
  Media Communications Lab\\
  University of Southern California\\
  Los Angeles, CA, USA \\
  \texttt{zazizi@usc.edu} \\
   \And
 Xuejing Lei \\
  Media Communications Lab\\
  University of Southern California\\
  Los Angeles, CA, USA \\
  \texttt{xuejing@usc.edu} \\
  \And
  C.-C. Jay Kuo \\
  Media Communications Lab\\
  University of Southern California\\
  Los Angeles, CA, USA \\
  \texttt{cckuo@sipi.usc.edu} \\
}

\begin{document}
\maketitle

\begin{abstract}
A simple and effective low-light image enhancement method based on a noise-aware texture-preserving retinex model is proposed in this work. The new method, called NATLE, attempts to strike a balance between noise removal and natural texture preservation through a low-complexity solution.  Its cost function includes an estimated piece-wise smooth illumination map and a noise-free texture-preserving reflectance map. Afterwards, illumination is adjusted to form the enhanced image together with the reflectance map. Extensive experiments are conducted on common low-light image enhancement datasets to demonstrate the superior performance of NATLE. 
\end{abstract}

% keywords can be removed

\section{Introduction}\label{sec:introduction}
Images captured in low-light conditions suffer from low visibility, lost
details and high noise.  Besides unpleasant human visual experience,
low-light images can significantly degrade the performance of computer
vision tasks such as object detection and recognition. As a result,
low-light enhancement is widely demanded in consumer electronics and
computer vision systems.  Quite a few low light enhancement methods have
been proposed. Early work is based on histogram equalization.  Although
being simple and fast, it suffers from unnatural color, amplified noise,
and under/over-exposed areas.  One popular approach for low-light
enhancement is based on retinex decomposition. In this work, we propose
to decompose the retinex model into the element-wise product of a
piece-wise smooth illumination map and a noise-aware texture-preserving
refelectance map, solve it with a simple and low complexity solution
and, finally, use gamma correction to enhance the illumination map.  We
give the new method an acronym, ``NATLE", due to its noise-aware
texture-preserving characteristics. 

Our main idea is sketched below. We begin with an initialization of a
piece-wise smooth illumination map.  Based on the retinex model and the
initial illumination map, we can solve for the initial refelectance map.
Afterwards, we apply nonlinear median filtering as well as linear
filtering \cite{fastABF} to the RGB channels of initial refelectance map
separately for noise-free, texture-preserving reflectance estimation.
The complexity of our solution is low.  Our work has several
contributions. First, it conducts low light enhancement and denoising at
the same time.  Second, it has a low-complexity solution. Third, it
preserves details without unrealistic edges for better visual quality. 

The rest of this paper is organized as follows. Related work is reviewed
in Sec.  \ref{sec:review}. Our new method is detailed in Sec.
\ref{sec:method}.  Experimental results are shown in Sec.
\ref{sec:experiments}.  Finally, concluding remarks and future work are
given in Sec.  \ref{sec:conclusion}.

\section{Related Work}\label{sec:review}

Motivated by the human vision system (HVS), retinex-based methods
decompose an image into an element-wise product of a reflectance map and
an illumination map and, then, adjust these two maps to enhance a
low-light image.  LR3M \cite{LR3M}, which is a state-of-the-art
retinex-based method, adopts an optimization framework that determines a
piece-wise smooth illumination map and a noise-free contrast-enhanced
reflectance map, denoted by $L$ and $R$, respectively, for a low-light
image. Although it yields a noise-free contrast-enhanced image, it
suffers from unrealistic bold borders surrounded with white halo on
edges (see Fig. \ref{fig:lr3m}). Moreover, its run time is longer (see
Table \ref{tab:score}). STAR \cite{STAR} is another low-light
enhancement method based on the retinex model. It finds $R$ and $L$
using structure and texture maps extracted from the input image.
Although STAR preserves texture well, it does not remove noise
effectively. The Generative Adversarial Networks (GANs) offer another
family of learning-based methods for low light enhancement. Examples
include RDGAN \cite{RDGAN} and EnlightenGAN \cite{EnlightenGAN}.  RDGAN
preserves texture well. Yet, its enhanced images tend to be noisy and
with faded-color.  EnlightenGAN can handle over-exposure without paired
data, it tends to have noisy results as pointed out in \cite{Zero}.
Another CNN-based method, known as Zero-DCE \cite{Zero}, estimates
enhancement curves for each pixel in an image.

\section{Proposed NATLE Method}\label{sec:method}

\begin{figure*}[t]
    \begin{subfigure}[t]{0.16\textwidth}
        \centering
        \includegraphics[width=\linewidth]{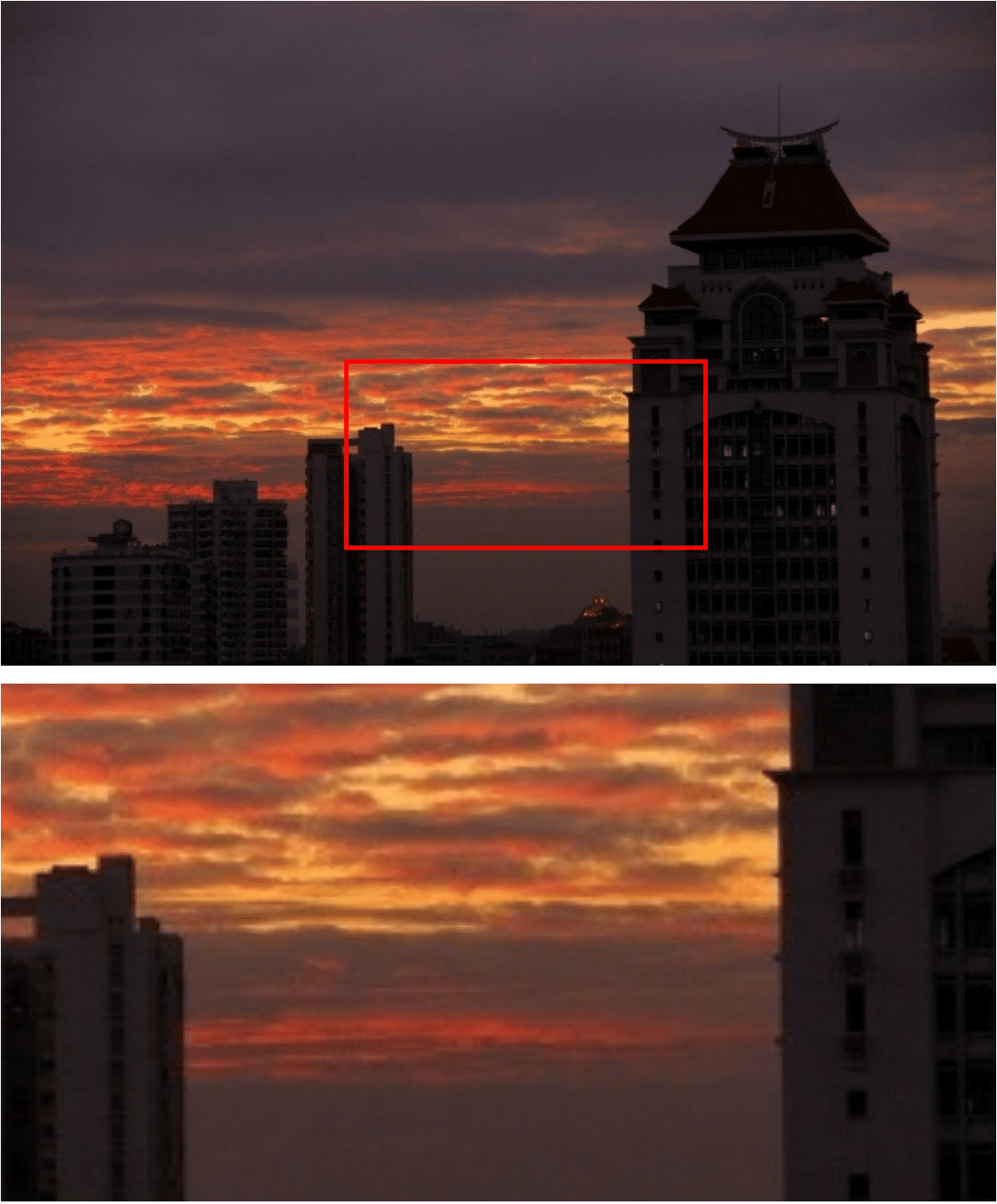} 
        \caption{input} \label{fig:in}
    \end{subfigure}
    \hfill
    \begin{subfigure}[t]{0.16\textwidth}
        \centering
        \includegraphics[width=\linewidth]{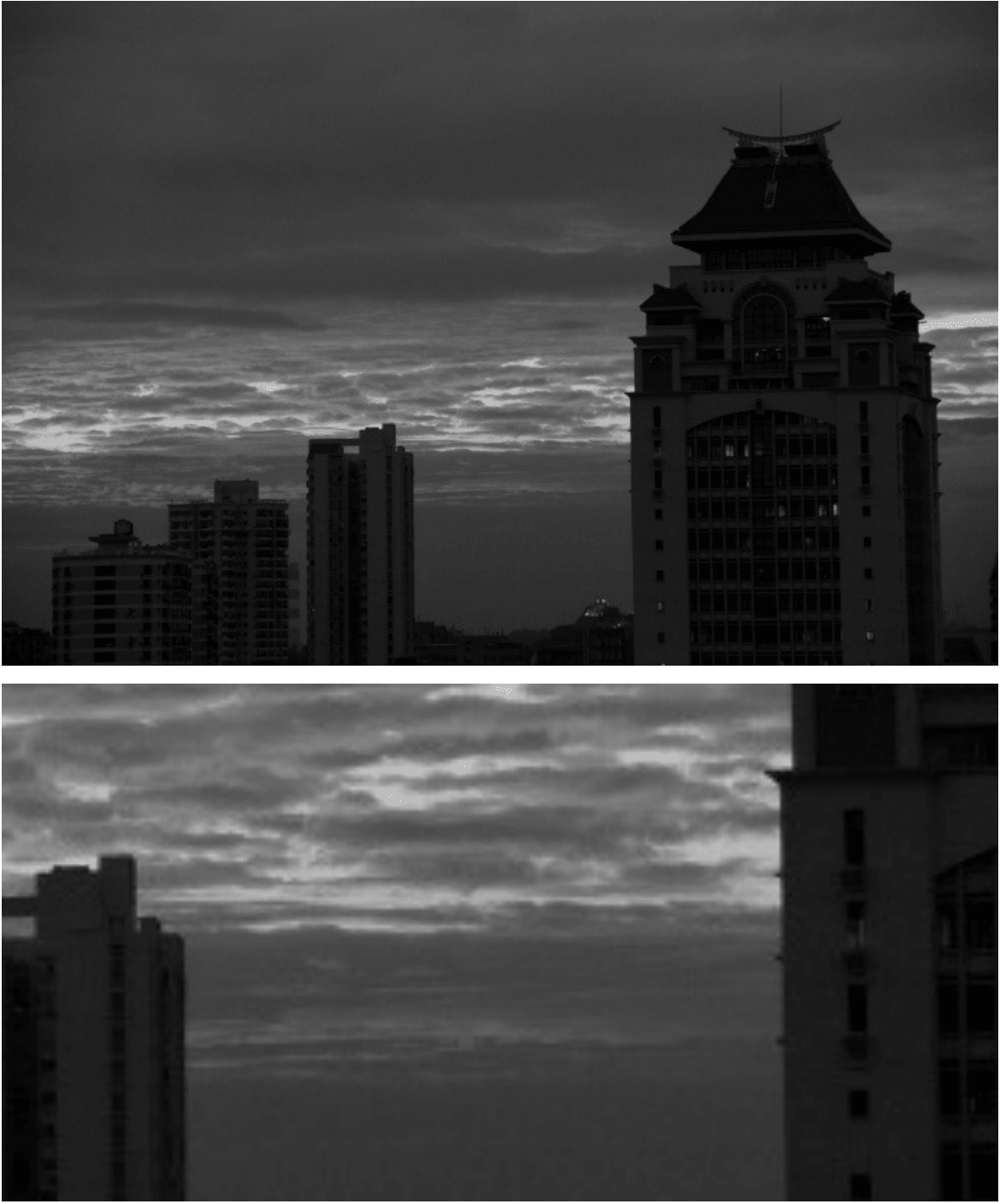} 
        \caption{$\widehat{L}$} \label{fig:lhat}
    \end{subfigure}
    \hfill
    \begin{subfigure}[t]{0.16\textwidth}
        \centering
        \includegraphics[width=\linewidth]{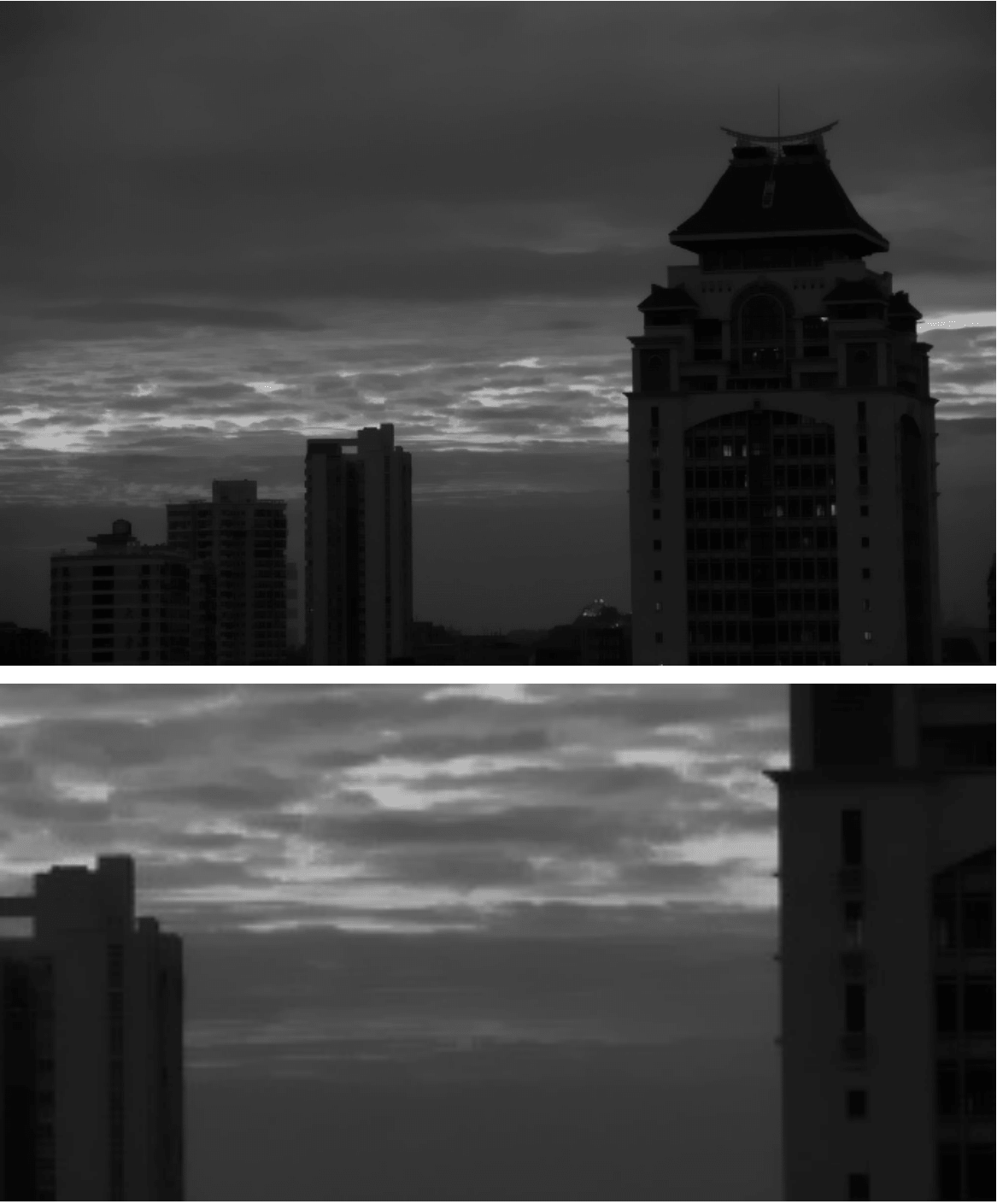} 
        \caption{L} \label{fig:l}
    \end{subfigure}
    \hfill
    \begin{subfigure}[t]{0.16\textwidth}
        \centering
        \includegraphics[width=\linewidth]{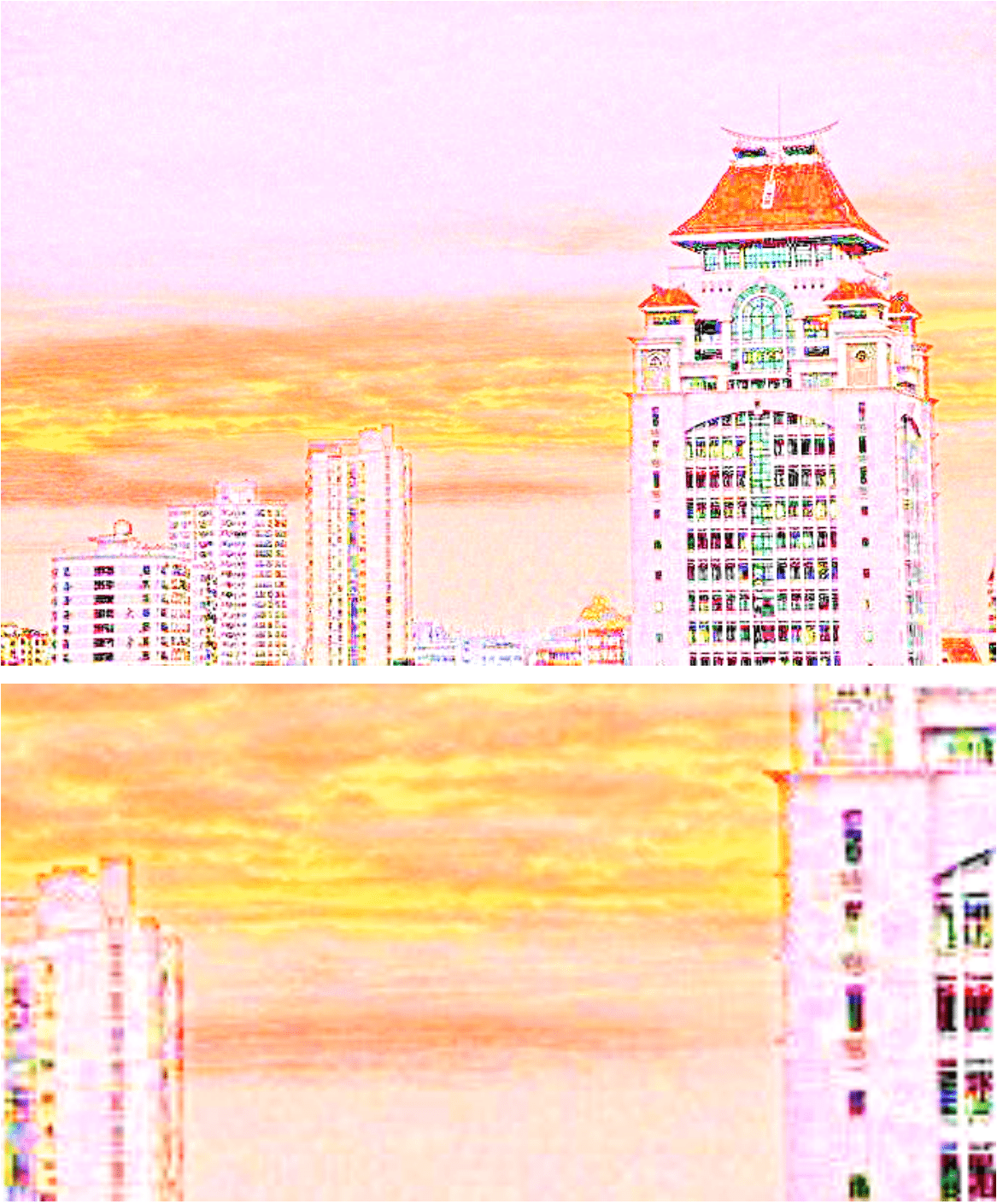} 
        \caption{Noisy $\widehat{R}$} \label{fig:rhat}
    \end{subfigure}
    \hfill
      \begin{subfigure}[t]{0.16\textwidth}
          \centering
         \includegraphics[width=\linewidth]{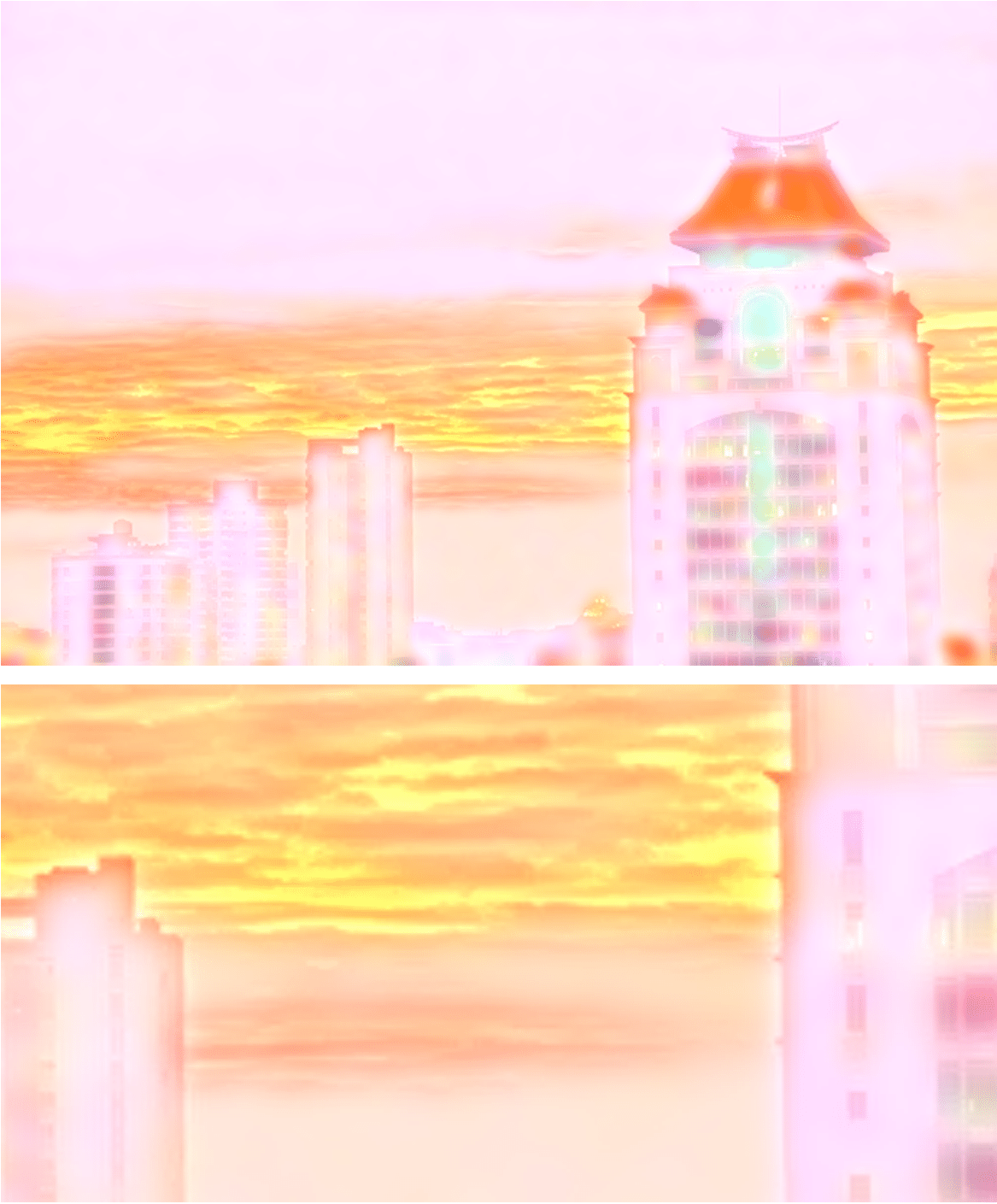} 
          \caption{R} \label{r}
      \end{subfigure}
      \hfill
     \begin{subfigure}[t]{0.16\textwidth}
         \centering
         \includegraphics[width=\linewidth]{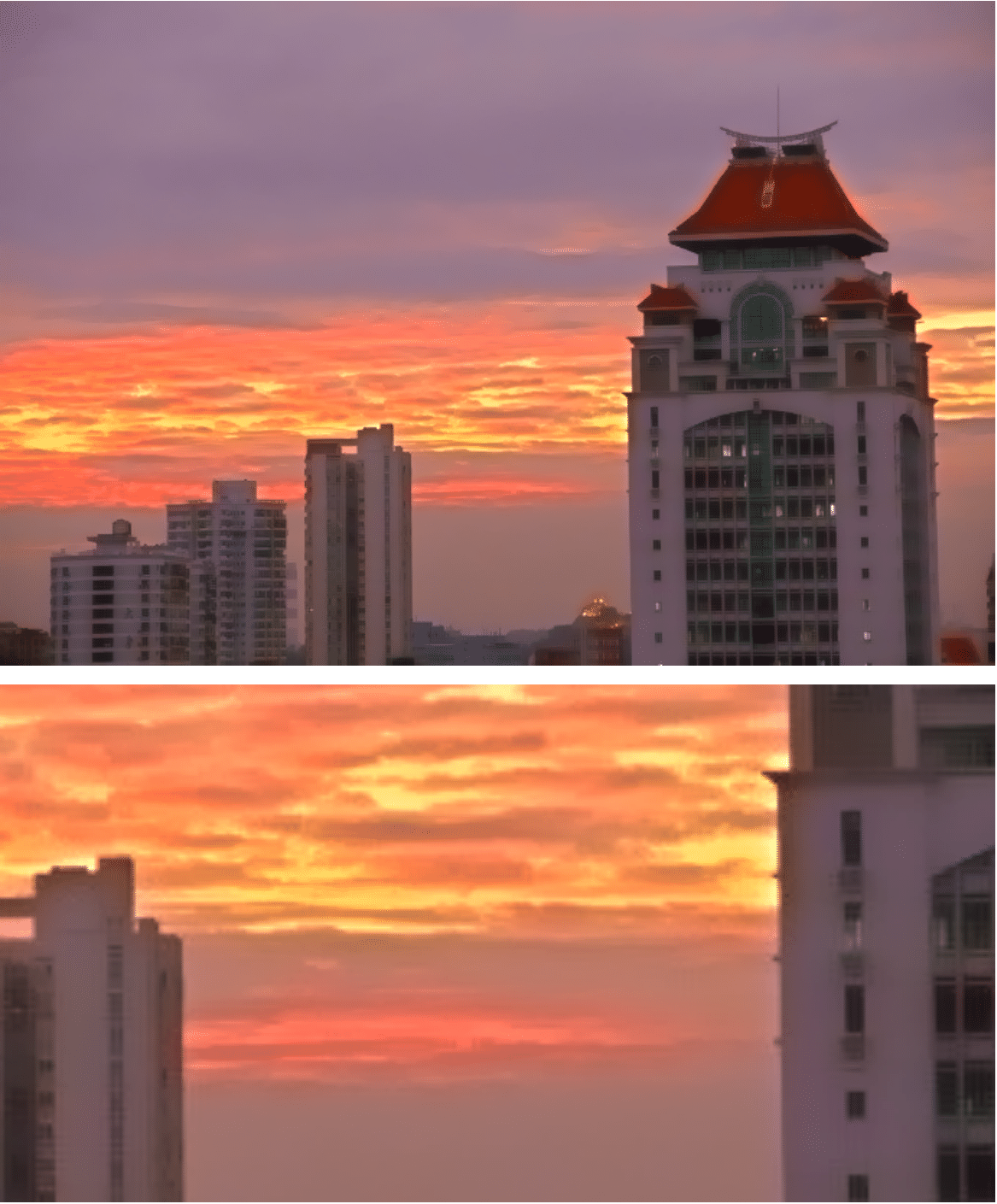} 
         \caption{output} \label{fig:out}
    \end{subfigure}
\caption{Illustration of NATLE's processing steps: (a) input, (b) illumination
map initialization, (c) illumination map estimation, (d) noisy
reflectance map, (e) reflectance map estimation, (f) output after
illumination gamma correction.} \label{fig:steps}
\end{figure*}

{\bf System Overview.} The classic retinex model decomposes an observed
image (S) into an element-wise multiplication of its reflectance map (R)
and its illumination map (L) as
\begin{equation}\label{eq:retinex}
S = R \circ L,
\end{equation}
where $R$ represents inherent features of the image, which is decoupled
from lightness, and $L$ delineates the lightness condition. A desired
refelectance map includes texture and details while an ideal
illumination map is a piece-wise smooth map indicative of the edge
information. The NATLE method consists of two major steps:
\begin{itemize}
\item Step 1: Use the first optimization procedure to estimate $L$; 
\item Step 2: Use the second optimization procedure to estimate $R$
based on estimated $L$ from Step 1. 
\end{itemize}
Then, with gamma correction, NATLE yields the final output image.  NATLE
is summarized in Algorithm \ref{alg1}. Its intermediate processing
results are illustrated in Fig.  \ref{fig:steps}. 

{\bf Step 1.} To estimate $L$, we conduct the following optimization \cite{LR3M}:
\begin{equation}\label{eq:L}
\argmin_L \|L-\widehat{L}\|_{F}^2+\alpha\|\nabla L\|_{1},
\end{equation}
where $\alpha$ is a model parameter and $\widehat{L}$ is an initial
estimation of $L$. It is set to the default average of RGB three color
components as $\widehat{L} = 0.299R+0.587G+0.114B$.  The first term in
the right-hand-side of Eq. (\ref{eq:L}) demands that $L$ represents the
luminance while the second term ensures that $L$ is a peice-wise smooth
map containing remarkable edges only. We have the following
approximation for the second term:
\begin{equation}\label{eq:Lappr}
\lim_{\epsilon \to 0^+}\sum_{x}\sum_{d\in\{h,v\}}\dfrac{(\nabla_{d}
L(x))^2}{\mid\nabla_{d}\widehat{L}(x)\mid+\epsilon}=\|\nabla L\|_{1},
\end{equation}
where $d$ is the gradient direction and $v$ and $h$ indicate the
vertical and horizontal directions, respectively.  Thus, Eq.
(\ref{eq:L}) can be rewritten as
\begin{equation}\label{eq:L2}
\argmin_L \|L-\widehat{L}\|_{F}^2+\sum_{x}\sum_{d\in\{h,v\}}A_{d}(x)(\nabla_{d} L(x))^2,
\end{equation}
where 
\begin{equation}
A_{d}(x)= \frac{\alpha}{\mid\nabla_{d}\widehat{L}(x)\mid+\epsilon}.
\end{equation}

%%%%%%%%%%%%%%%%%%%%%%%%%%%%%%%%%%%%%%%%%%%%%%%%%%%%
%%%%%%%%%%%%%%%%%%%%%%%%%%%%%%%%%%%%%%%%%%%%%%%%%%%%

{\bf Step 2.} The reflectance map, $R$, can be solved with the
following optimization:
\begin{equation}\label{eq:R}
\argmin_R \|R-\widehat{R}\|_{F}^2+\beta\|\nabla R-G\|_{F}^2,   
\end{equation}
where $\widehat{R}$ is a noise-free initialization of $R$:
\begin{equation}\label{eq:Rhat}
\widehat{R}=S\oslash(L+\varepsilon)-N,
\end{equation}

where $S$ is the V channel of the input in the HSV color space, $N$ is
input noise, $L$ is the estimated illumination obtained in Step 1,
$\oslash$ denotes the element-wise division and $\varepsilon$ is a small
value to prevent division by zero. 

\begin{figure*}[t]
    \begin{subfigure}[t]{0.16\textwidth}
        \centering
        \includegraphics[width=\linewidth,height=4.71\linewidth]{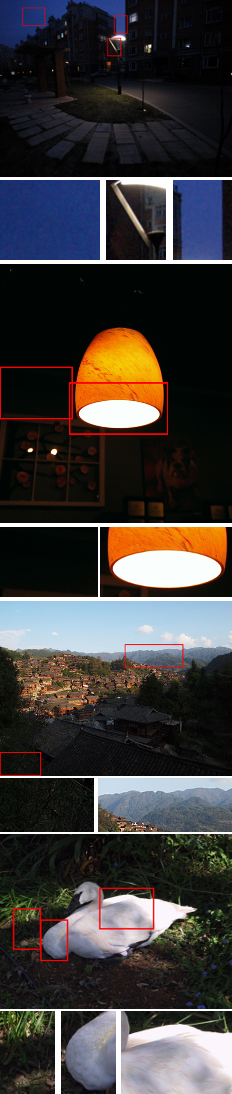} 
        \caption{Input} \label{fig:input}
    \end{subfigure}
    \hfill
    \begin{subfigure}[t]{0.16\textwidth}
        \centering
        \includegraphics[width=\linewidth,height=4.71\linewidth]{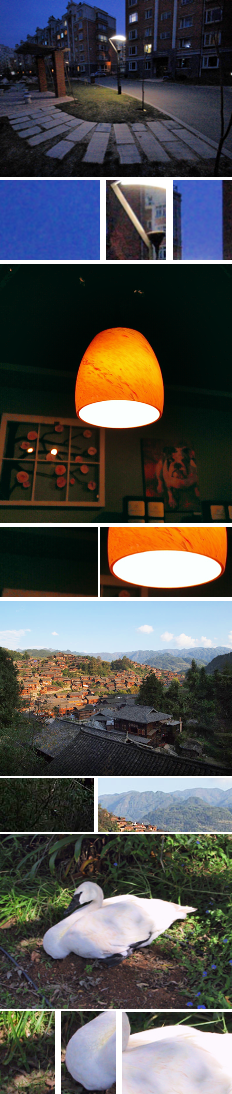} 
        \caption{PIE \cite{PIE}} \label{fig:pie}
    \end{subfigure}
    \hfill
    \begin{subfigure}[t]{0.16\textwidth}
        \centering
        \includegraphics[width=\linewidth,height=4.71\linewidth]{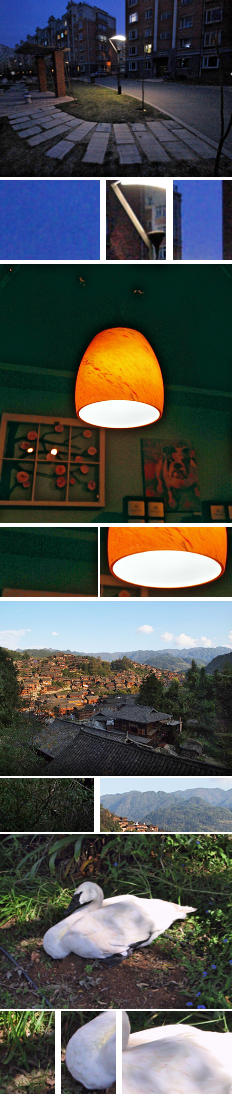} 
        \caption{SRIE \cite{SRIE}} \label{fig:srie}
    \end{subfigure}
    \hfill
      \begin{subfigure}[t]{0.16\textwidth}
          \centering
         \includegraphics[width=\linewidth,height=4.71\linewidth]{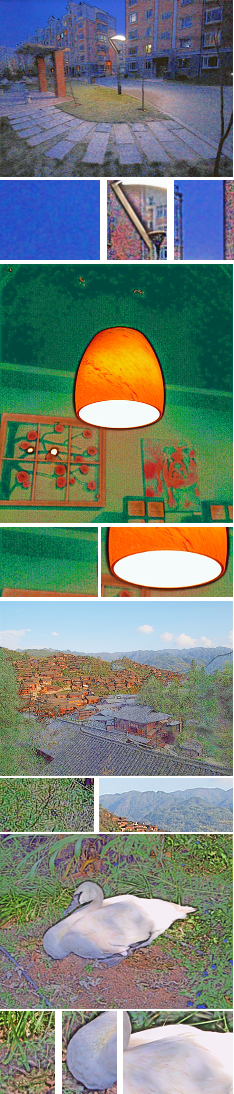} 
          \caption{RetinexNet \cite{retinexnet}} \label{fig:retinexnet}
      \end{subfigure}
    \hfill
    \begin{subfigure}[t]{0.16\textwidth}
        \centering
        \includegraphics[width=\linewidth,height=4.71\linewidth]{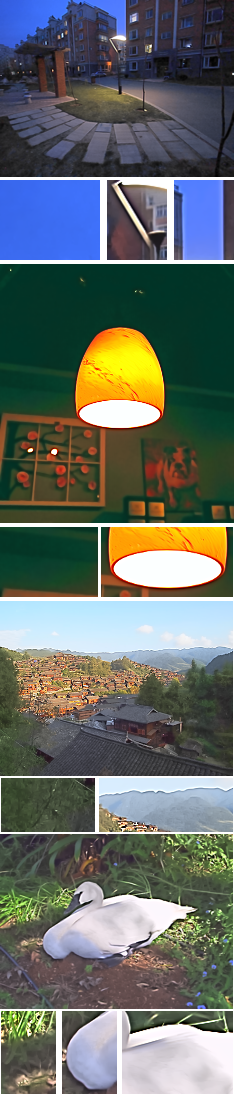} 
        \caption{LR3M \cite{LR3M}} \label{fig:lr3m}
    \end{subfigure}
      \hfill
    \begin{subfigure}[t]{0.16\textwidth}
        \centering
        \includegraphics[width=\linewidth,height=4.71\linewidth]{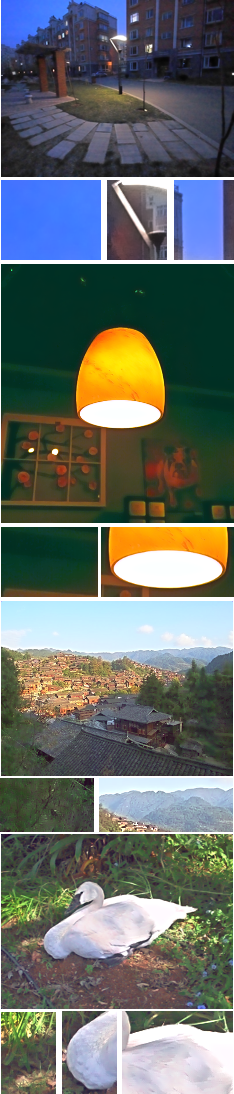} 
        \caption{NATLE (Ours)} \label{fig:ours}
    \end{subfigure}
\caption{Qualitative comparison of low-light enhancement results.}\label{fig:quality1}
\end{figure*}

The first term in Eq. (\ref{eq:R}) is used to ensure: 1) the
element-wise multiplication of $R$ and $L$ gives $S$, as demanded by the
retinex model; 2) $R$ is noise-free since $\widehat{R}$ is the
element-wise division of $S$ by $L$ minus noise.  Being inspired by
\cite{LR3M}, the second term in Eq. (\ref{eq:R}) has texture-preserving
and noise-removal dual roles in our model.  It enforces $R$ not to
involve very small gradients to reduce noise. Yet, it is modified in our
model to avoid bold borders between objects or halo next to edges, which
appear in \cite{LR3M}. 
When $\nabla S$ is small, \cite{LR3M} attempts to enhance the contrast by forcing $\nabla R$
to be much larger than $\nabla S$; however, it also
affects edges that should not be amplified much, leading to unrealistic
borders or halo near edges. In our model, $\nabla R$ is forced to be
slightly greater than $\nabla S$ everywhere, except for very small
gradients. This results in natural borders without halo as shown in Fig.
\ref{fig:quality1}. To this end, we modify $G$ in \cite{LR3M} as

\vspace{0.25cm}
\begin{equation}\label{eq:G}
G=\begin{cases}
   0, \quad \quad  \nabla S<\epsilon_{g} \\
   \lambda\nabla S,
   \end{cases}
\end{equation}

where $\epsilon_{g}$ is the threshold to filter out small gradients and
$\lambda$ controls the degree of amplification. 

To obtain noise-free $R$, input noise $N$ should be removed as shown in
Eq. (\ref{eq:Rhat}). To do so, we convert $S\oslash(L+\varepsilon)$, hue
and saturation to the RGB color space and conduct denoising there.
First, a median filter is applied to RGB channels to remove color
impulse noise.  Second, the Fast Adaptive Bilateral Filtering (fastABF)
method \cite{fastABF} is used to remove remaining noise in each of RGB
channels. FastABF employs the Gaussian kernel of different parameters at
different pixel locations to adjust denoising degree according to the
local noise level. Furthermore, it can implemented by a fast algorithm.
After denoising, the RGB output is converted back to the HSV space.  Hue
and saturation are saved for the final solution. The V channel serves as
$\widehat{R}$ to initialize $R$. The above procedure not only remove
noise in the $V$ channel of noisy $\widehat{R}$ but also in hue and
saturation channels. 

It is worthwhile to point out that conventional high-performance
denoising methods such as non-local-mean (NLM) \cite{NLM} and BM3D
\cite{BM3D} are slow in run time and weak in texture preservation.
Faster denoising methods such as classic bilateral filtering
\cite{bilateral} do not work well for images with heavy noise.  The
fastABF method is a good solution since it provides a good balance
between low complexity, texture preservation and effective noise
reduction.

{\bf Closed-Form Solution.} The optimization problems in Eqs.
(\ref{eq:L}) and (\ref{eq:L2}) can be solved by differentiating with
respect to $L$ and $R$ and setting the derivative to zero in a
straightforward manner. There is no approximation neither
high-complexity algorithm in order to solve them. Actually, the final
solution can be derived in closed form as
\begin{eqnarray}
l&=&(I+\sum_{d\in\{h,v\}}{D_{d}^T {\mbox Diag} (a_{d})D_{d}})^{-1}\widehat{l}, \label{eq:l} \\
r&=&(I+\beta\sum_{d\in\{h,v\}}{D_{d}^2})^{-1}(\widehat{r}+\beta\sum_{d\in\{h,v\}}
{D_{d}^T g_{d}}), \label{eq:r}
\end{eqnarray}
where $\mbox{Diag} (y)$ is a diagonal matrix of vectorized $y$ from
matrix $Y$, $D_{d}$ is a discrete differential operator matrix that
plays the role of $\nabla$ and $I$ is the identity matrix.  Once
vectors $l$ and $r$ are determined, they are reformed to matrices $L$
and $R$, respectively.  As the last step, gamma correction is applied to
$L$ to adjust illumination. The ultimate enhanced low-light image 
can be computed as
\begin{equation}\label{eq:Sprime}
S^{'} = R \circ L^{\frac{1}{\gamma}}. 
\end{equation}

%%%%%%%%%%%%%%%%%%%%%%%%%%%%%%%%%%%%%%%%%%%%%%%%%%%%%%
\begin{figure}[t]
  \centering
  \begin{minipage}{.6\linewidth}
\begin{algorithm}[H]
%\SetAlgoLined
  \caption{Low-Light Enhancement Algorithm}
  \label{alg1}
  \textbf{Input}: Low-light Image
  \begin{algorithmic}[1]
    \State Illumination initialization;
    \State Illumination estimation via Eq. \eqref{eq:l};
    \State Reflectance initialization via Eq. \eqref{eq:Rhat};
      \begin{itemize}
          \item $S$ is the $V$ channel of input in the HSV space;
          \item Noisy $\widehat{R}$ with element-wise division (S/L);
          \item Noisy $\widehat{R}$ back to the RGB space;
          \item Apply median filtering and FastABF to RGB 3 channels;
          \item Return to the HSV space:
          \begin{itemize}
              \item $\widehat{R}$ is the V channel;
              \item Save denoised hue and saturation;
          \end{itemize}
      \end{itemize}
    \State Reflectance estimation via Eq. \eqref{eq:r};
    \State $S^{'}$ is enhanced $S$ via Eq. \eqref{eq:Sprime};
    \State Integrate $S^{'}$, hue \& saturation and change to the RGB space
    \end{algorithmic}
    \textbf{Output}: Normal-light Image
\end{algorithm}
\end{minipage}
\end{figure}
%%%%%%%%%%%%%%%%%%%%%%%%%%%%%%%%%%%%%%%%%%%%%%%%%%%%%%

%%%%%%%%%%%%%%%%%%%%%%%%%%%%%%%%%%%%%%%%%%%%%%%%%%%%%%
%%%%%%%%%%%%%%%%%%%%%%%%%%%%%%%%%%%%%%%%%%%%%%%%%%%%%%

%%%%%%%%%%%%%%%%%%%%%%%%%%%%%%%%%%%%%%%%%%%%%%
\begin{figure*}[htb]
    \begin{subfigure}[t]{0.24\textwidth}
        \centering
        \includegraphics[width=\linewidth]{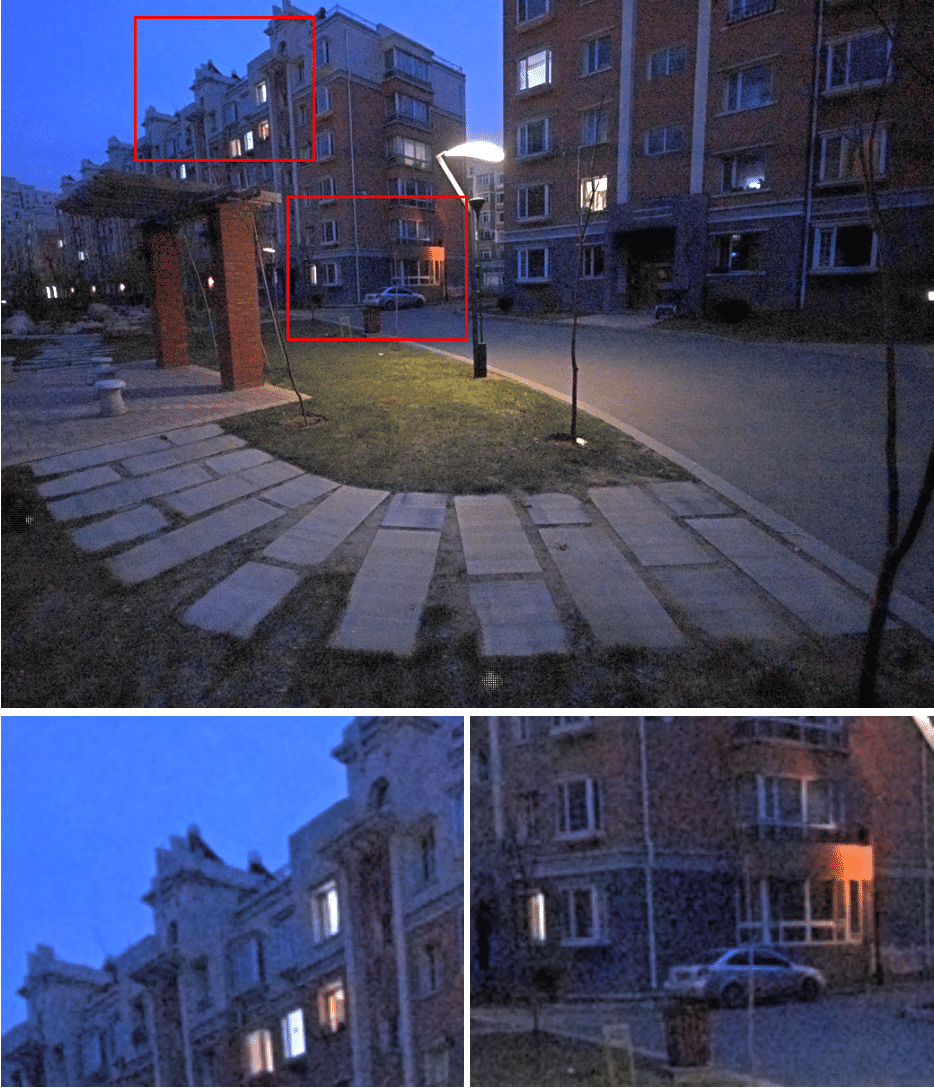} 
        \caption{$\alpha=0$} \label{fig:alpha0}
    \end{subfigure}
    \hfill
    \begin{subfigure}[t]{0.24\textwidth}
        \centering
        \includegraphics[width=\linewidth]{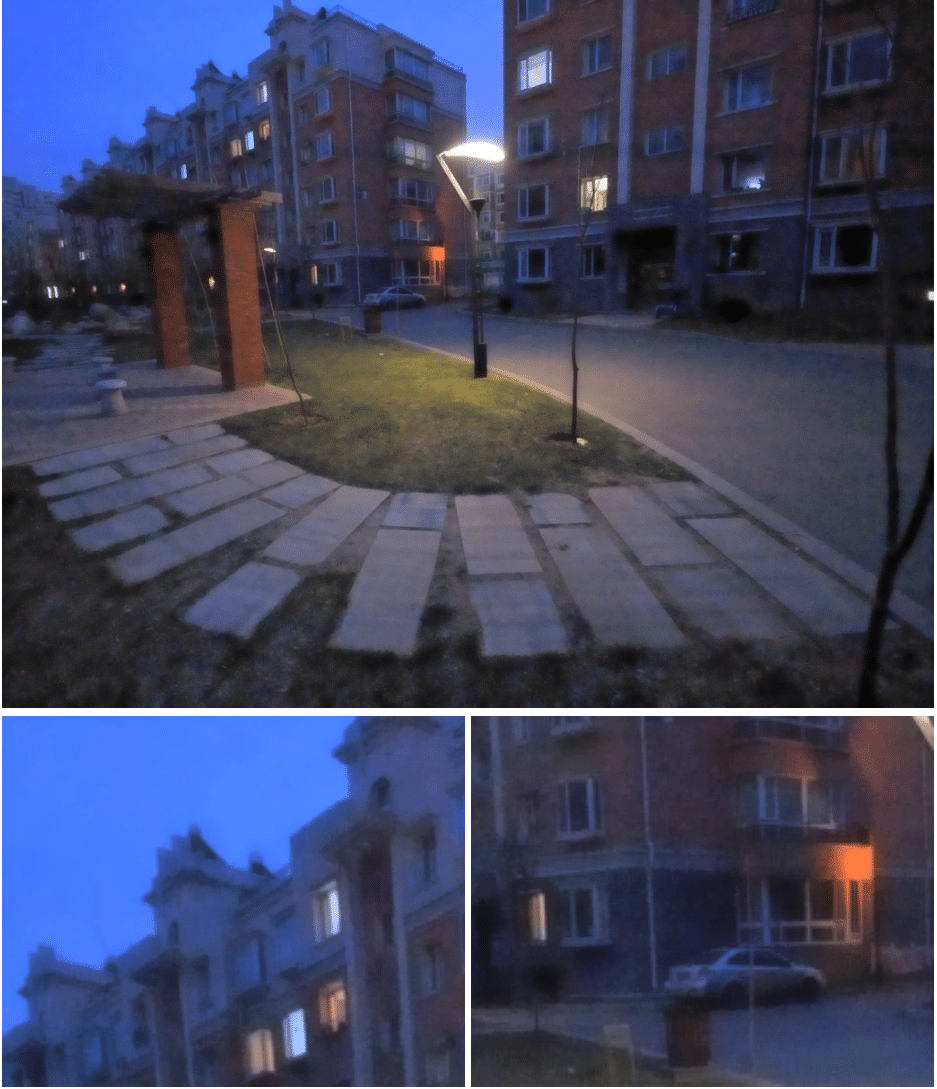} 
        \caption{$\beta=0$} \label{fig:beta0}
    \end{subfigure}
    \hfill
    \begin{subfigure}[t]{0.24\textwidth}
        \centering
        \includegraphics[width=\linewidth]{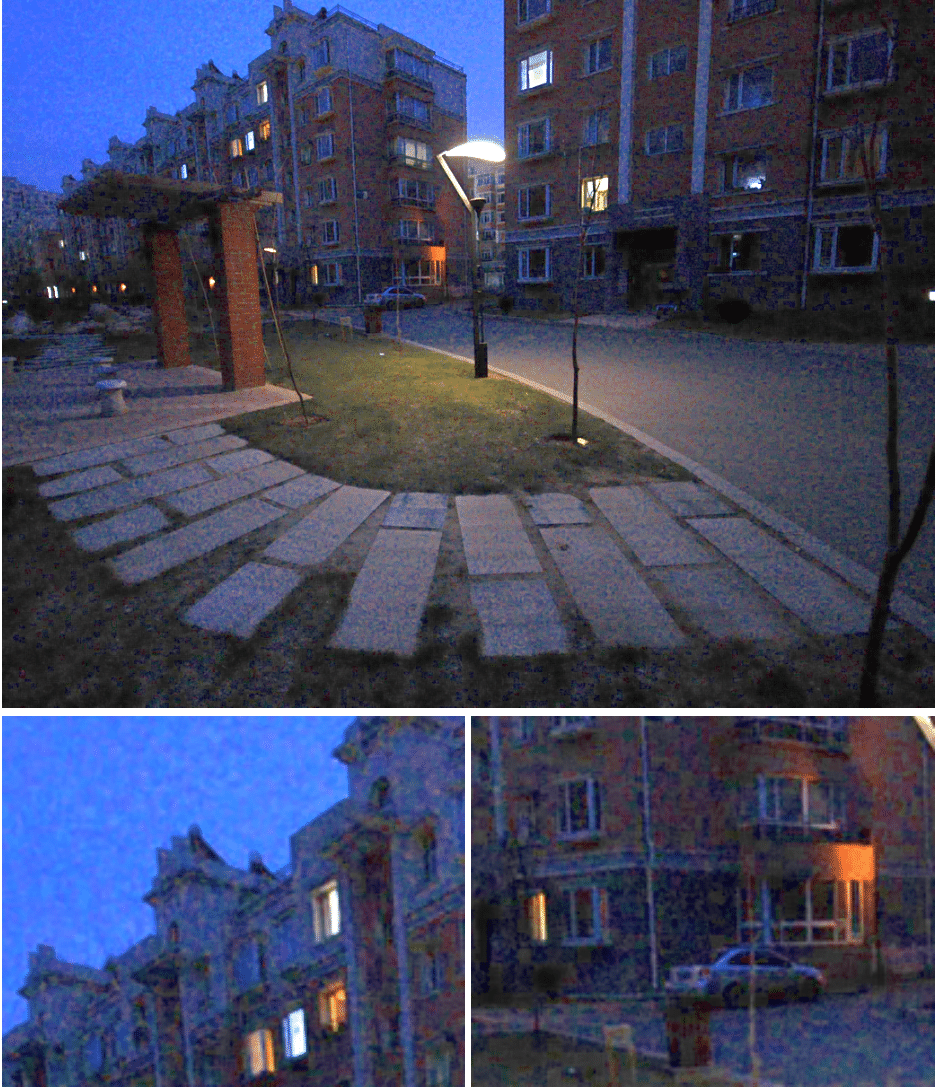} 
        \caption{Without denoising $\widehat{R}$} \label{fig:noisy}
    \end{subfigure}
    \hfill
    \begin{subfigure}[t]{0.24\textwidth}
        \centering
        \includegraphics[width=\linewidth]{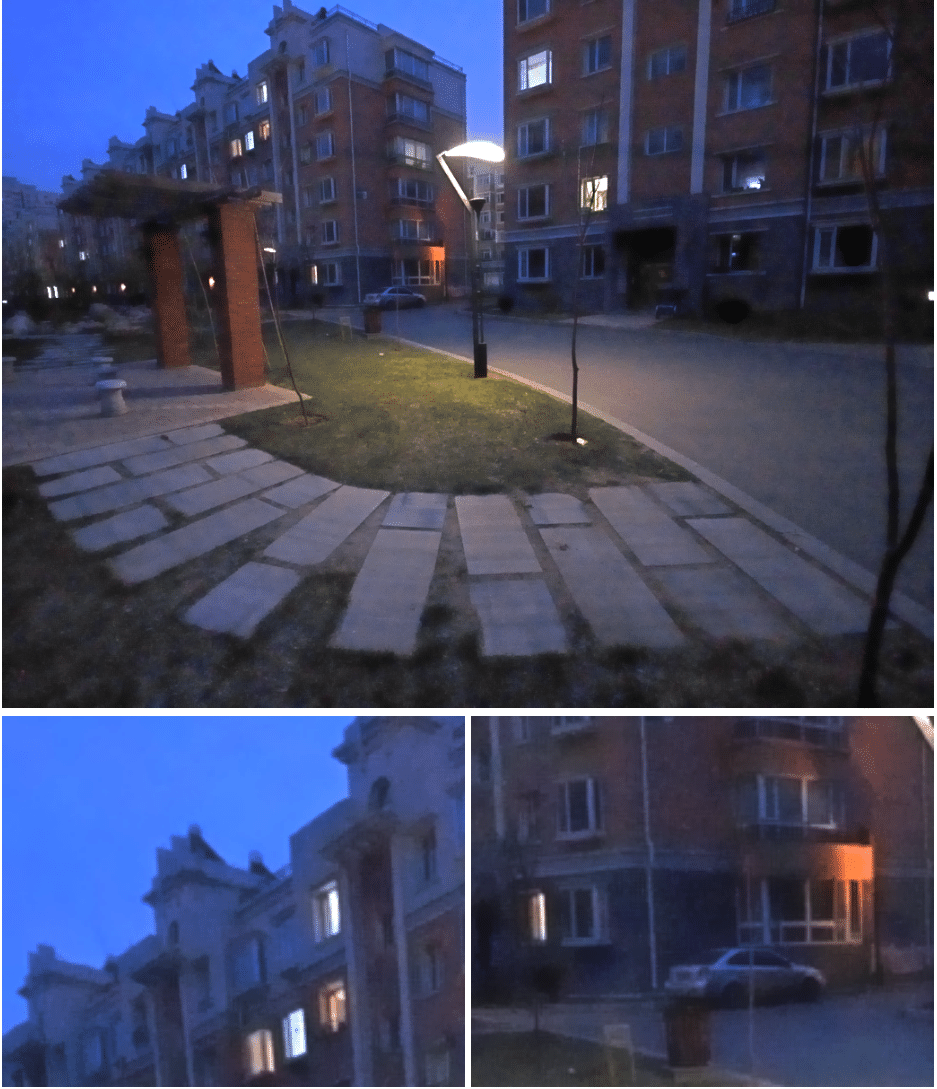} 
        \caption{NATLE (Ours)} \label{fig:withall}
    \end{subfigure}
\caption{Parameter study for NATLE: (a) result with $\alpha=0$, (b)
result with $\beta=0$, (c) result without denoising $\widehat{R}$, (d)
desired parameters.} \label{fig:parameters}
\end{figure*}
%%%%%%%%%%%%%%%%%%%%%%%%%%%%%%%%%%%%%%%%%%%%%%

\section{Experiments}\label{sec:experiments}

We conduct experiments on Matlab R2019b with an Intel Core i7 CPU
@2.7GHz. The parameters $\alpha$, $\beta$, $\lambda$ and $\gamma$ are
set to 0.015, 3, 1.1 and 2.2, respectively. We carry out performance
comparison of several benchmarking methods with two sets of images: 1)
59 commonly used low-light images collected from various datasets for
subjective evaluation and no-reference comparison; 2) 15 images from the
LOL paired dataset for reference-based comparison. 

{\bf Objective Evaluation.} A no-reference metric, ARISMC \cite{ARISMC},
and a reference-based one, SSIM \cite{SSIM}, are adopted for objective
evaluation of several methods in Table \ref{tab:score}. While ARISMC
assesses quality of both luminance and chrominance, SSIM evaluates
structural similarity with the ground truth. The run time is also
compared in the table. LR3M has the best ARISMC performance. NATLE has
the best SSIM performance and the second best ARISMC. Yet, LR3M demands
25 more times than NATLE. 

%%%%%%%%%%%%%%%%%%%%%%%%%%%%%%%%%%%%%%%%%%%%%%%%%
\begin{table}[h]
\caption{Objective performance evaluation}\label{tab:score}
\centering
\begin{tabular}{c|c|c|c}
\hline
{\color[HTML]{343434} Method} & ARISMC $\downarrow$          & SSIM $\uparrow$      & Run Time (sec.) $\downarrow$  \\ \hline
BIMEF \cite{BIMEF}            & 3.1543          & 0.5903          & \textbf{0.27}           \\
CRM \cite{CRM}                & 3.1294          & 0.5366          & 0.32                    \\
MF \cite{MF}                  & 3.1342          & 0.4910          & 0.96                    \\
PIE \cite{PIE}                & 3.0636          & 0.5050          & 1.55                    \\
SRIE \cite{SRIE}              & 3.1416          & 0.4913          & 16.89                   \\
LR3M \cite{LR3M}              & \textbf{2.7262} & 0.4390          & 127                     \\
NATLE (Ours)                  & 2.9970          & \textbf{0.6193} & 4.98                    \\ \hline
\end{tabular}
\end{table}
%%%%%%%%%%%%%%%%%%%%%%%%%%%%%%%%%%%%%%%%%%%%%%%%%

{\bf Subjective Evaluation.} A qualitative comparison of our method with
four benchmarking methods is shown in Fig. \ref{fig:quality1}. For the
first street image, PIE, SRIE and RetinexNet amplify noise in the
low-light enhanced image.  RetinexNet has unnatural color. LR3M has
extra borders or halo next to edges. For the second lamp image, PIE and
SRIE have either dark or noisy background. RetinexNet is noisy and
over-exposed with unnatural texture and color. LR3M has false red
borders around the lamp. For the third hills image, PIE and SRIE has
low-light results. RetinexNet reveals square traces of BM3D denoising on
trees and has unnatural color. LR3M removes all texture in mountains and
generates an extra border between mountain and sky. For the last bird
image, PIE and SRIE have low-light shadow areas.  RetinexNet and LR3M
generate black border around the bird. RetinexNet has unnatural color
while LR3M loses feather texture and blurs shadow area. NATLE yields
noise-free images with natural edges in these examples. It enhances
light adequately and preserves texture well. 

{\bf Discussion.} It is worthwhile to highlight several characteristics of the proposed NATLE method. 

{\em a) Denoising and Texture Preservation.} NATLE effectively removes
noise without losing texture detail when being applied to a wide range
of low-light images. The optimization in Eq. \eqref{eq:R} demands the
enhanced reflectance map to be as close as its noise-free form while
preserving edges and textures in the input. As compared with
\cite{LR3M}, NATLE takes a moderate approach. That is, it does not
denoise $R$ more than needed. This is the main reason why NATLE can
preserve texture, remove noise and maintain natural borders without
halo at the same time. 

{\em b) Speed.} NATLE performs fast and efficient with a closed-form
solution. It is of low-complexity, since it does not demand iterations.
It does not require sequential mathematical approximations, either. The
performance of NATLE is affected by denoising methods. FastABF is chosen
here. It is fine to adopt other denoising methods depending on the
application requirement. 

{\em c) Parameter Study.} The impact of model parameters $\alpha$,
$\beta$, $\lambda$ and $\gamma$ is shown in Fig. \ref{fig:parameters}. The results of setting $\alpha=0$ is shown in column (a).  It removes
the second term in Eq. \eqref{eq:L}, leading to a non-smooth
illumination map and a noisy output.  Moreover, removing this term
results in color distortions such as the yellow area on grass beside the
pavement at the bottom of the image. Column (b) shows results with
$\beta=0$.  Without the second term in Eq. \eqref{eq:R}, edges and
details are blurred. Column (c) is very noisy. It shows the need of
denoising $\widehat{R}$. Column (d) is the result by including all
model parameters, which is clearly better than the other three cases. 

\section{Conclusion and Future Work}\label{sec:conclusion}

A low-light image enhancement method based on a noise-aware
texture-preserving retinex model, called NATLE, was proposed in this
work.  It has closed-form solutions to the two formulated optimization
problems and allows fast computation. Its superior performance was
demonstrated by extensive experiments with both objective and subjective
evaluations. One possible future work is to extend this framework into
video low-light enhancement. The main challenge is to preserve temporal
consistency of enhanced video. 

\bibliographystyle{unsrt}  
\bibliography{references}  %%% Remove comment to use the external .bib 

\end{document}